\documentclass[conference]{IEEEtran}

\IEEEoverridecommandlockouts
\usepackage{algorithm}
\usepackage{algpseudocode}
\usepackage{cite}
\usepackage{amsmath,amssymb,amsfonts}

\usepackage{graphicx}
\usepackage{textcomp}
\usepackage{xcolor}
\usepackage{booktabs}
\usepackage{multirow}
\usepackage{array}

\def\BibTeX{{\rm B\kern-.05em
    {\sc i\kern-.025em b}\kern-.08em
    T\kern-.1667em\lower.7ex\hbox{E}\kern-.125emX}}

\begin{document}

\title{Mission-Level Runtime Assurance Framework for Autonomous Driving}
\author{
\IEEEauthorblockN{Chieh Tsai}
\IEEEauthorblockA{
Department of Electrical and Computer Engineering\\
The University of Arizona\\
Tucson, Arizona, USA\\
vegetableclean@arizona.edu
}
\and
\IEEEauthorblockN{Salim Hariri}
\IEEEauthorblockA{
Department of Electrical and Computer Engineering\\
The University of Arizona\\
Tucson, Arizona, USA\\
hariri@arizona.edu
}
}

\maketitle

\begin{abstract}

This paper studies runtime assurance for autonomous driving under faulty high-level autonomy commands.
Existing runtime-safety approaches mainly focus on platform-level safety, but a command may remain locally safe while still becoming mission-infeasible by skipping checkpoints, entering restricted regions, or exhausting the remaining mission budget. To address this limitation, we propose a mission-level runtime admissibility framework that jointly evaluates platform safety and future mission feasibility before command execution.
The framework introduces a successor-safe runtime monitor that rejects commands predicted to violate platform or mission admissibility under future successor evolution. Experimental results show that platform-level runtime safety alone does not detect mission-level inadmissibility, while the proposed framework improves mission success from about 0.44 to 0.73 under randomized mission-fault conditions.

\end{abstract}

\section{Introduction}

Autonomous driving systems increasingly rely on learning-enabled planning and control to operate in complex environments.
Most existing runtime-safety approaches focus on platform-level safety, such as collision avoidance, lane stabilization, and low-level vehicle recovery~\cite{seto1998simplex,chen2022runtime}.
Although these methods are effective for preventing immediate unsafe behavior, they do not necessarily determine whether a high-level autonomy command can still successfully complete the mission.

In practice, an autonomy command may remain locally safe while still causing mission failure.
For example, a planner may skip a required checkpoint, enter a restricted region, or choose a route that later becomes infeasible under the remaining mission budget.
Such failures may not immediately produce collisions, but they still violate the operational objective of the autonomous system.

Existing runtime-assurance and safety-filtering approaches primarily evaluate local driving safety and often do not reason about future mission feasibility.
As a result, platform-safe but mission-infeasible autonomy behavior may still be accepted during execution.

These limitations motivate the need for mission-level runtime admissibility validation.
High-level autonomy outputs should be validated against both platform-level safety and future mission feasibility before execution.
Hierarchy alone is insufficient because mission-infeasible behavior may still propagate through the autonomy stack even when local vehicle behavior remains collision-free.

To address this problem, this paper proposes a hierarchical mission-level runtime-assurance framework for autonomous driving under faulty high-level autonomy commands.
The framework separates mission-level admissibility evaluation from platform-level safety monitoring and introduces a successor-safe runtime monitor that performs predictive successor-based admissibility evaluation before execution. Commands predicted to violate platform safety or mission feasibility are rejected and replaced by a fallback action.

The proposed framework further extends highway-env~\cite{leurent2018environment} with mission-command fault injection inspired by prior autonomous-vehicle fault-injection methodology~\cite{jha2019ml,jha2019kayotee,schmedding2024strategic}.
The benchmark includes checkpoint-skipping faults, restricted-region violations, and future mission-infeasibility scenarios for evaluating mission-level runtime admissibility.

The main contributions of this work are summarized as follows:

\begin{itemize}

\item We introduce a mission-level runtime-assurance framework capable of rejecting faulty high-level autonomy commands before execution.

\item We propose a successor-safe runtime monitor that jointly evaluates platform safety and future mission feasibility.

\item We develop a mission-command fault-injection benchmark on top of Highway-envelop for evaluating mission-level autonomy failures.

\item We compare the proposed framework against platform-level runtime-safety baselines under randomized mission-fault conditions.

\end{itemize}
\section{Related Work}

\subsection{Runtime Assurance and Safety Filtering}

Runtime assurance provides a common architecture for deploying advanced or learning-enabled controllers with runtime monitoring and fallback control.
The Simplex architecture formalized the use of a high-performance controller together with a verified baseline controller that takes over when safety conditions are threatened~\cite{seto1998simplex}.
This idea has been extended to autonomous systems and autonomous driving through runtime safety filtering and Simplex-style driving architectures~\cite{schierman2020runtime,hobbs2023runtime,chen2022runtime,peng2023rta}.
Related maneuver-verification and trajectory-safety methods check future collision risk, reachable occupancy, road-boundary safety, and fallback feasibility before execution~\cite{althoff2014online,van2011reciprocal}.

These works motivate the use of runtime filtering before autonomy outputs are executed.
However, their primary focus is platform-level safety, such as collision avoidance, drivable-area compliance, or local maneuver feasibility.
In contrast, this work evaluates whether a high-level command preserves future mission feasibility, including checkpoint completion, restricted-region avoidance, and remaining-budget feasibility.

\subsection{Mission-Level Planning and Runtime Monitoring}

Mission-level autonomy has been studied through temporal-logic planning, execution monitoring, and task-and-motion planning frameworks.
Prior work has shown how high-level temporal objectives can be represented, monitored, and connected to lower-level motion generation in robotic and autonomous systems~\cite{doherty2009temporal,zudaire2021assumption,bonnah2022runtime,pereira2023task}.
These methods establish the importance of tracking mission progress beyond immediate vehicle safety.

The present work differs in its runtime assurance role.
Rather than synthesizing a complete mission plan, we validate candidate high-level commands before execution.
The goal is to detect commands that remain locally safe but become mission-inadmissible under future successor evolution.
These methods typically assume formal mission-specification or planning interfaces different from the runtime command-validation setting considered in this work.

\subsection{Fault Injection for Autonomous-Vehicle Resilience}

Fault injection is widely used to evaluate autonomous-vehicle resilience under abnormal operating conditions.
DriveFI and Kayotee study AV fault injection across software, hardware, and system components~\cite{jha2019ml,jha2019kayotee}.
Other studies examine sensor failures, actuator faults, and strategic neural-network fault injection in safety-critical scenarios~\cite{matos2024survey,holzmann2023fault,schmedding2024strategic}.

Our benchmark follows the same resilience-evaluation philosophy but changes the injection target.
Instead of injecting faults into sensors, perception modules, or low-level controllers, we inject faults into high-level mission commands.
This exposes a failure mode that platform-safety monitors may miss: commands that remain locally safe while making the remaining mission infeasible.

\section{Proposed Method}

This section presents the proposed hierarchical mission-level runtime-assurance framework for autonomous driving under faulty high-level autonomy commands.
The objective of the framework is to extend conventional platform-level runtime safety into mission-level runtime admissibility by validating candidate autonomy commands before execution.

\begin{figure*}[t]
\centering
\includegraphics[width=\textwidth]{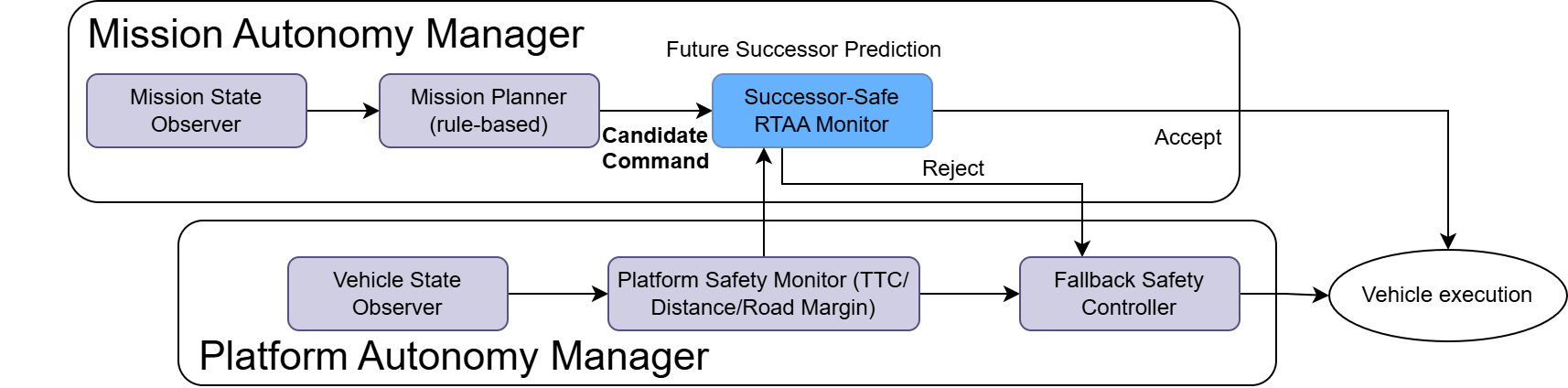}
\caption{Proposed hierarchical mission-level runtime-assurance architecture. The mission layer evaluates candidate commands using successor-safe mission admissibility prediction, while the platform layer maintains low-level driving safety and fallback recovery.}
\label{fig:rtaa_architecture}
\end{figure*}

Fig.~\ref{fig:rtaa_architecture} illustrates the overall architecture of the proposed framework.
The system is organized into two interacting layers:
a mission-autonomy layer responsible for high-level mission admissibility evaluation and a platform-autonomy layer responsible for low-level driving safety and fallback recovery.

The mission-autonomy layer first observes mission-related states including checkpoint progress, restricted-region status, and remaining mission budget.
A mission planner then generates a candidate high-level command.
The proposed successor-safe RTAA monitor evaluates whether the predicted successor evolution of the command remains admissible under both platform-safety and mission-feasibility constraints before execution.

The platform-autonomy layer continuously monitors low-level driving safety metrics including collision distance, time-to-collision (TTC), and road-margin safety.
If the RTAA monitor rejects the candidate command, the platform layer activates a fallback safety controller to preserve safe vehicle operation.

Unlike conventional runtime-safety approaches that primarily evaluate local driving safety only after unsafe behavior emerges, the proposed framework validates future mission admissibility before command execution.
This allows the framework to reject commands that remain locally platform-safe but progressively become mission-infeasible.

The remainder of this section introduces the mission-level runtime-admissibility formulation, the successor-safe runtime monitor, and the mission-command fault-injection benchmark used for evaluation.

\subsection{Mission-Level Runtime Admissibility}

We consider an autonomous vehicle operating under high-level commands generated by a mission planner.
The mission specification is defined as:
\begin{equation}
\varphi=(C,G,R,T),
\end{equation}
where $C$ denotes required checkpoints, $G$ denotes the goal region, $R$ denotes restricted regions, and $T$ denotes the remaining mission fuel budget.

As shown in Fig.~\ref{fig:rtaa_architecture}, the mission-autonomy layer contains three main components:
the mission-state observer, the mission planner, and the successor-safe RTAA monitor.

The mission-state observer continuously tracks mission-related states including checkpoint completion, restricted-region proximity, traffic context, and remaining fuel budget.
Based on the observed mission state, the mission planner generates a candidate high-level command for vehicle execution.

Platform safety is defined using standard driving-safety constraints:
\begin{equation}
\mathcal{S}_{p}
=
\{x
\mid
d_{\mathrm{col}}\ge d_{\min},
\ \mathrm{TTC}\ge\tau_{\min},
\ m_{\mathrm{road}}\ge m_{\min}
\}.
\end{equation}

Mission feasibility requires that the remaining mission can still be completed while satisfying checkpoint, restricted-region, and budget constraints:
\begin{equation}
(x_t,z_t)\in\mathcal{S}_{m},
\end{equation}
where $x_t$ denotes the platform state and $z_t$ denotes the mission-progress state.

The joint admissible region is defined as:
\begin{equation}
\mathcal{A}=\mathcal{S}_{p}\cap\mathcal{S}_{m}.
\end{equation}

A command is considered runtime-admissible if the predicted successor rollout remains inside the admissible region over a finite horizon:
\begin{equation}
(\hat{x}_{t+k},\hat{z}_{t+k})\in\mathcal{A},
\quad
k=1,\ldots,H.
\end{equation}

\begin{table}[t]
\caption{Runtime admissibility semantics.}
\label{tab:admissibility}
\centering
\small
\begin{tabular}{ccc}
\toprule
Platform Safety & Mission Feasibility & Decision \\
\midrule
Safe & Feasible & Accept \\
Safe & Infeasible & Reject \\
Unsafe & Feasible & Reject \\
Unsafe & Infeasible & Reject \\
\bottomrule
\end{tabular}
\end{table}

Table~\ref{tab:admissibility} summarizes the runtime decision semantics.
The key distinction of the proposed framework is the detection of commands that remain platform-safe but become mission-infeasible under future successor evolution.

\subsection{Successor-Safe Runtime Monitor}

At each decision step, the mission planner generates a nominal command:
\begin{equation}
u_t^{\mathrm{nom}}.
\end{equation}

The successor-safe RTAA monitor predicts future platform and mission evolution over a finite horizon:
\begin{equation}
(\hat{x}_{t+k},\hat{z}_{t+k}),
\quad
k=1,\ldots,H.
\end{equation}

The monitor jointly evaluates predicted platform safety, checkpoint progress, restricted-region avoidance, and remaining mission feasibility before execution.

If any predicted successor violates platform safety or mission feasibility, the nominal command is rejected and replaced by a fallback action:
\begin{equation}
u_t=
\begin{cases}
u_t^{\mathrm{nom}},
&
u_t^{\mathrm{nom}}\in\mathcal{U}_{\mathrm{adm}},
\\
u_t^{\mathrm{safe}},
&
\text{otherwise}.
\end{cases}
\end{equation}

The runtime-assurance procedure is summarized in Algorithm~\ref{alg:rtaa}.

\begin{algorithm}[t]
\caption{Successor-Safe Runtime Assurance}
\label{alg:rtaa}
\footnotesize
\begin{algorithmic}[1]

\Require Current state $(x_t,z_t)$, nominal command $u_t^{\mathrm{nom}}$, horizon $H$

\State $\hat{x}_{t}\leftarrow x_t$
\State $\hat{z}_{t}\leftarrow z_t$

\For{$k=1$ to $H$}

\State Predict successor platform state:
\[
\hat{x}_{t+k}\leftarrow
f(\hat{x}_{t+k-1},u_t^{\mathrm{nom}})
\]

\State Update mission-progress state:
\[
\hat{z}_{t+k}\leftarrow
g(\hat{z}_{t+k-1},\hat{x}_{t+k})
\]

\If{$\hat{x}_{t+k}\notin\mathcal{S}_{p}$
\textbf{or}
$(\hat{x}_{t+k},\hat{z}_{t+k})\notin\mathcal{S}_{m}$}

\State Reject nominal command
\State Execute fallback action
\State \Return $u_t^{\mathrm{safe}}$

\EndIf

\EndFor

\State Accept nominal command
\State \Return $u_t^{\mathrm{nom}}$

\end{algorithmic}
\end{algorithm}

Unlike conventional runtime-safety approaches that primarily evaluate local driving safety, the proposed framework evaluates both platform safety and future mission feasibility before command execution.

The proposed framework can operate with rule-based planners, optimization-based planners, or learning-enabled controllers.
In the current implementation, the mission planner is rule-based, while the RTAA monitor acts as a runtime admissibility layer compatible with multiple command-generation modules.
In the current implementation, the fallback controller performs conservative lane-following and speed-reduction maneuvers to preserve platform safety while attempting to maintain mission recoverability.
\subsection{Mission-Command Fault Injection}

The evaluation benchmark extends highway-env~\cite{leurent2018environment} with mission-command fault injection inspired by prior autonomous-vehicle fault-injection methodology~\cite{jha2019ml,jha2019kayotee,schmedding2024strategic}.

Each injected fault is represented as:
\begin{equation}
F=(\tau,a,d,s,h),
\end{equation}
where $\tau$ denotes the fault type, $a$ the activation condition, $d$ the duration, $s$ the severity level, and $h$ the expected downstream hazard.

Unlike prior autonomous-vehicle fault injection that targets sensors, perception modules, or low-level controllers, this work injects faults into high-level mission commands.

The benchmark includes checkpoint-skipping faults, restricted-region traversal faults, and future mission-infeasibility faults.
Faults are strategically activated near checkpoint decision regions, restricted-zone boundaries, and mission-budget bottlenecks in order to evaluate whether the proposed runtime monitor can detect mission inadmissibility before recoverability is lost.

\begin{table*}[t]
\caption{Structured mission-command fault-injection taxonomy.}
\label{tab:mission_fault_taxonomy}
\centering
\begin{tabular}{p{0.14\textwidth}p{0.15\textwidth}p{0.15\textwidth}p{0.18\textwidth}p{0.24\textwidth}}
\toprule
Fault type & Activation region & Duration & Severity levels & Expected downstream hazard \\
\midrule
Checkpoint fault & Checkpoint decision region & Persistent or partial recovery & Mild delayed approach; moderate inefficient reroute; severe bypass & Required checkpoint is reached late or skipped, causing mission failure despite platform-safe driving. \\
Restricted-zone fault & Restricted-zone boundary & Persistent or partial recovery & Mild boundary grazing; moderate proximity; severe traversal & Vehicle approaches or enters a restricted mission region without necessarily causing collision. \\
Future-infeasibility fault & Mission-budget bottleneck & Persistent or partial recovery & Mild reduced time margin; moderate recoverable delay; severe unrecoverable delay & Locally safe progress becomes insufficient to complete checkpoint and goal within the remaining budget. \\
\bottomrule
\end{tabular}
\end{table*}

\section{Experimental Results}

This section evaluates whether the proposed mission-level runtime-assurance framework can preserve mission correctness under faulty high-level autonomy commands.
The experiments are designed to answer four main questions:
whether platform-level runtime safety alone is sufficient for mission-level autonomy,
whether successor-safe admissibility improves robustness under randomized mission faults,
how the successor horizon affects runtime behavior and false rejection,
and what factors contribute to the remaining RTAA failure cases.

The evaluation therefore focuses not only on final mission success, but also on mission-infeasibility detection, runtime intervention behavior, robustness under stochastic mission faults, false rejection, and fallback recovery limitations.

\subsection{Experimental Setup}

The proposed framework is evaluated in highway-env~\cite{leurent2018environment} augmented with mission constraints and high-level mission-command faults.
The benchmark includes required checkpoints, restricted mission regions, remaining mission-budget constraints, and successor-based mission-feasibility evaluation.

The experiments compare six runtime-autonomy configurations:
the faulty planner without runtime assurance,
a current-state mission monitor,
a platform-only future monitor,
an ORCA-based trajectory-safety adapter~\cite{van2011reciprocal},
a Simplex platform runtime-assurance baseline~\cite{chen2022runtime},
and the proposed successor-safe RTAA framework.

All methods operate under identical traffic distributions, mission layouts, restricted regions, and fault schedules.
The randomized evaluation uses seeds 0--2, three checkpoint layouts, and both persistent and partial-recovery mission-fault variants.
This produces 108 mission-fault rollouts per method.
Unless otherwise stated, reported uncertainty corresponds to 95\% confidence intervals across rollouts.

The evaluation records mission success, mission-infeasibility detection rate, restricted-region violations, fallback frequency, false rejection, runtime overhead, and recovery behavior.
The objective is not only to evaluate local driving safety, but also to determine whether runtime assurance can preserve mission feasibility under faulty autonomy behavior.

\subsection{Baseline Comparison}

We first compare the proposed RTAA framework against platform-level runtime-safety baselines and diagnostic runtime monitors.
The purpose of this experiment is to determine whether platform-level runtime safety alone is sufficient for handling mission-level autonomy faults.

\begin{table*}[t]
\caption{Baseline comparison on randomized mission-fault episodes. Each row reports mean $\pm$ 95\% confidence interval over rollout evaluations.}
\label{tab:open_source_baseline_comparison}
\centering
\begingroup
\scriptsize
\setlength{\tabcolsep}{2.4pt}

\begin{tabular}{llcccccc}
\toprule
Method & Type & Success & Detection & R-zone viol. & Fallbacks & Runtime/step \\
\midrule

No runtime assurance 
& Diagnostic 
& 0.44$\pm$0.09 
& 0.00$\pm$0.00 
& 0.31$\pm$0.09 
& 0.00$\pm$0.00 
& \textcolor{blue}{0.000 ms} \\

Current-state mission monitor 
& Diagnostic 
& 0.44$\pm$0.09 
& 0.41$\pm$0.09 
& 0.29$\pm$0.09 
& 30.33$\pm$10.68 
& \textcolor{blue}{0.003 ms} \\

Platform-only future monitor 
& Diagnostic 
& 0.48$\pm$0.09 
& 0.00$\pm$0.00 
& 0.31$\pm$0.09 
& 2.76$\pm$1.16 
& \textcolor{blue}{0.075 ms} \\

Open-source ORCA adapter~\cite{van2011orca}
& Open-source 
& 0.40$\pm$0.09 
& 0.00$\pm$0.00 
& 0.41$\pm$0.09 
& 49.63$\pm$9.31 
& 9.578 ms \\

Simplex platform RTA~\cite{chen2022runtime}
& Adapted 
& 0.44$\pm$0.09 
& 0.00$\pm$0.00 
& 0.31$\pm$0.09 
& 1.38$\pm$0.41 
& 12.237 ms \\

Proposed successor-safe RTAA 
& Proposed 
& \textcolor{red}{\textbf{0.73$\pm$0.08}} 
& \textcolor{red}{\textbf{0.59$\pm$0.09}} 
& \textcolor{red}{\textbf{0.15$\pm$0.07}} 
& 5.43$\pm$1.21 
& \textcolor{blue}{0.080 ms} \\

\bottomrule
\end{tabular}

\vspace{1mm}
\footnotesize
108 mission-fault rollouts per method using seeds 0--2 and three checkpoint layouts.

\endgroup
\end{table*}

\begin{figure*}[t]
\centering
\includegraphics[width=\textwidth]{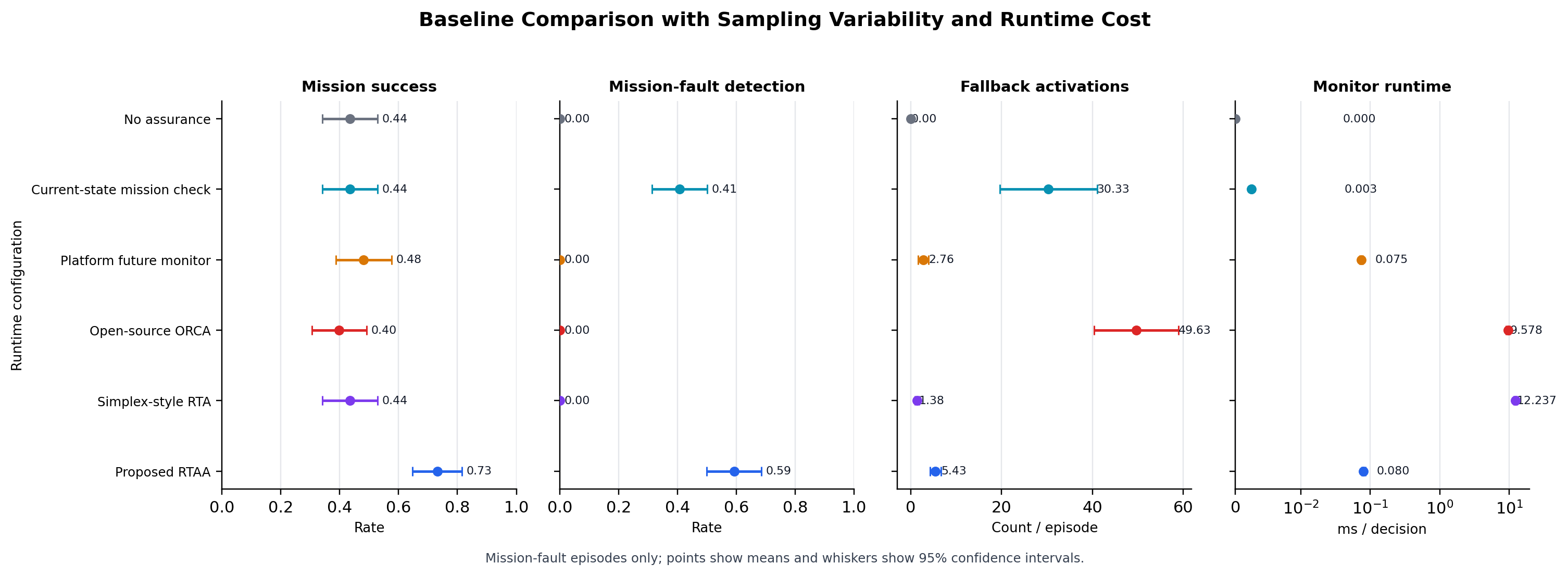}
\caption{Aggregate baseline comparison under randomized mission faults. Error bars denote 95\% confidence intervals across rollout evaluations.}
\label{fig:baseline_metric_point_ranges}
\end{figure*}

Table~\ref{tab:open_source_baseline_comparison} and Fig.~\ref{fig:baseline_metric_point_ranges} summarize the aggregate baseline comparison under randomized mission faults.
The results show that platform-level runtime safety alone does not detect mission-level inadmissibility.

The ORCA adapter and Simplex runtime-assurance baseline successfully respond to local driving risk such as collision avoidance and unsafe vehicle behavior, but they do not evaluate checkpoint completion, restricted-region constraints, or remaining mission budget.
As a result, both methods produce zero mission-infeasibility detection under the aggregate mission-fault benchmark.

The current-state mission monitor detects some mission-rule violations, but its intervention often occurs after mission feasibility has already degraded significantly.
In contrast, the proposed successor-safe RTAA framework predicts future mission evolution before execution and rejects commands whose predicted successors leave the admissible region.

The proposed RTAA improves mission success from approximately $0.44\pm0.09$ for the faulty planner and platform-level runtime baselines to $0.73\pm0.08$, while simultaneously achieving nonzero mission-infeasibility detection.
The runtime-overhead measurements further show that the proposed successor-based monitor remains computationally lightweight in the tested Python benchmark.

These results support the central claim of the paper:
platform-safe behavior does not necessarily imply mission-admissible behavior.

\subsection{Randomized Mission-Fault Robustness}

We next evaluate robustness under randomized mission-fault conditions.
The purpose of this experiment is to determine whether the proposed runtime monitor generalizes beyond deterministic mission scenarios.

The benchmark introduces randomized traffic density, stochastic fault activation timing, multiple checkpoint layouts, and partial-recovery fault conditions.
Traffic density is sampled from varying traffic configurations, while mission layouts include adjacent-early, opposite-late, and staggered-mid checkpoint arrangements.

\begin{figure*}[t]
\centering
\includegraphics[width=\textwidth]{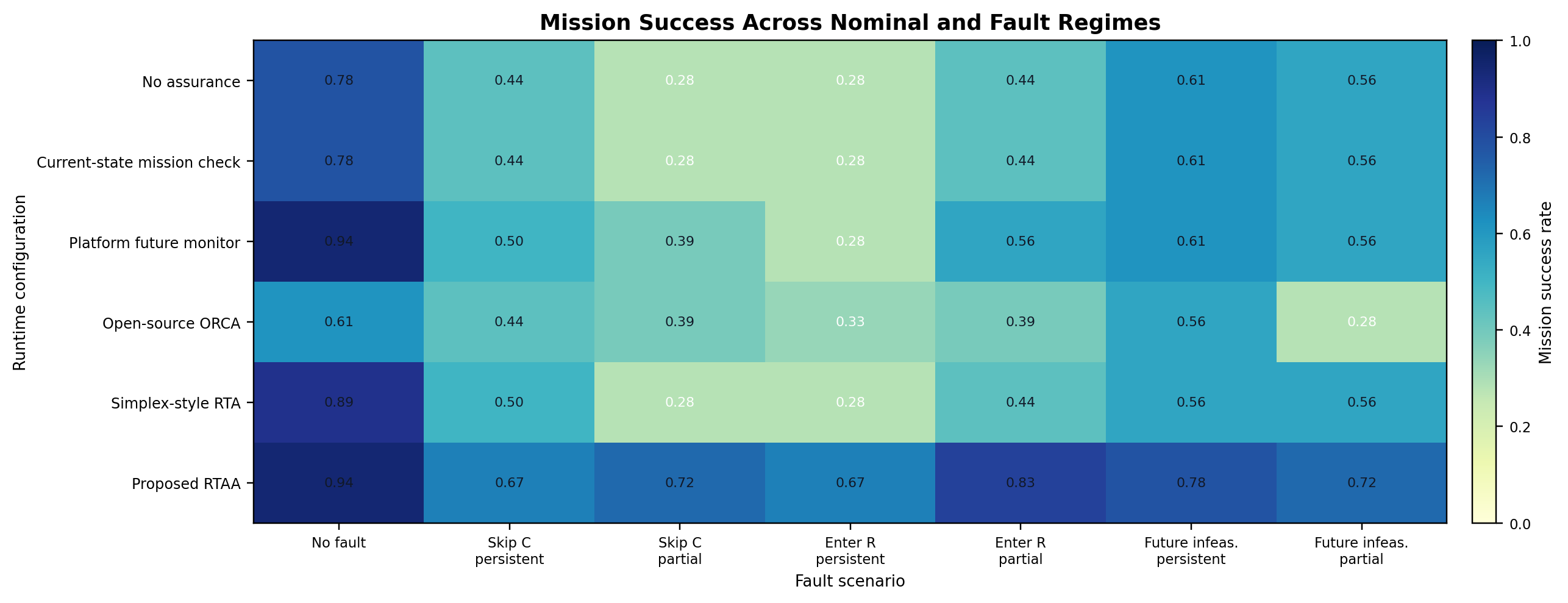}
\caption{Mission success under randomized mission-fault scenarios and checkpoint layouts.}
\label{fig:robust_mission_success}
\end{figure*}

Fig.~\ref{fig:robust_mission_success} shows mission success under the randomized mission-fault benchmark.
The results demonstrate that RTAA consistently improves mission success relative to the faulty planner and platform-level runtime-safety baselines across different mission layouts and fault conditions.

The observed improvement originates from successor-safe admissibility evaluation.
Instead of reacting only to immediate driving risk, the proposed monitor predicts future mission evolution and rejects commands whose successors leave the joint admissible region:
\begin{equation}
\mathcal{A}=\mathcal{S}_{p}\cap\mathcal{S}_{m}.
\end{equation}

The randomized-fault evaluation therefore supports the main hypothesis of the paper:
platform-level runtime safety alone is insufficient for preserving mission-level correctness under faulty high-level autonomy commands.
\subsection{Learning-Enabled Controller Pilot}

To verify that the result is not tied to a purely rule-based command source, we add a small learning-enabled controller pilot.
The bundled pretrained Stable-Baselines3 DQN highway policy is used as an advanced local command source without retraining.
Because the available DQN was trained on grayscale highway observations and discrete meta-actions, its output is adapted into the same mission-command interface by mapping lane-left, lane-right, faster, slower, and idle actions to target lane and speed updates.
The low-level continuous controller and RTAA monitor are otherwise unchanged.
This pilot is intentionally small: it uses 36 randomized mission-fault rollouts per method and reports only mission success, mission-fault detection, restricted-zone violation, and fallback count.

\begin{table}[t]
\caption{Learning-enabled controller pilot using a pretrained DQN command source without RL retraining.}
\label{tab:learning_controller_pilot}
\centering
\begingroup
\scriptsize
\setlength{\tabcolsep}{2pt}
\begin{tabular}{lccccc}
\toprule
Method & $n$ & Succ. & Detect. & R-viol. & Fb. \\
\midrule
Rule + prop. RTAA 
& 36 
& 0.75$\pm$0.14 
& 0.69$\pm$0.15 
& 0.17$\pm$0.12 
& 5.67$\pm$1.62 \\

DQN only 
& 36 
& 0.28$\pm$0.15 
& 0.00$\pm$0.00 
& 0.22$\pm$0.14 
& 0.00$\pm$0.00 \\

DQN + plat. RTA 
& 36 
& 0.25$\pm$0.14 
& 0.00$\pm$0.00 
& 0.22$\pm$0.14 
& 1.53$\pm$0.75 \\

DQN + prop. RTAA 
& 36 
& \textbf{0.75$\pm$0.14} 
& \textbf{0.89$\pm$0.10} 
& \textbf{0.17$\pm$0.12} 
& 7.33$\pm$2.48 \\
\bottomrule
\end{tabular}
\endgroup
\end{table}

Table~\ref{tab:learning_controller_pilot} shows that the pretrained DQN command source alone does not solve the mission-fault benchmark.
Adding a platform-only Simplex shield constrains local driving risk but still does not detect mission inadmissibility.
When the same DQN command source is wrapped by the proposed mission-level RTAA, mission success increases from $0.28\pm0.15$ to $0.75\pm0.14$, while mission-fault detection rises from zero to $0.89\pm0.10$.
Thus, the learning-enabled pilot supports the same conclusion as the main comparison: the key contribution is mission-level runtime admissibility validation, not the particular planner or controller used to generate the candidate command.

\subsection{Horizon and False-Rejection Analysis}

We further analyze the effect of successor-prediction horizon on runtime-assurance behavior.
The purpose of this experiment is to evaluate the tradeoff between early mission-infeasibility detection and conservative runtime intervention.

The successor horizon is varied over:
\begin{equation}
H\in\{1,2,4,6,8\}.
\end{equation}

\begin{table}[t]
\caption{RTAA successor-horizon ablation under randomized mission faults.}
\label{tab:rtaa_horizon_ablation}
\centering
\begin{tabular}{cccccc}
\toprule
$H$ & Success & Future detect & R-zone viol. & Fallbacks & Cost \\
\midrule
1 & 0.52 & 0.97 & 0.15 & 16.77 & 1.38 \\
2 & 0.54 & 0.94 & 0.15 & 15.36 & 1.08 \\
4 & 0.72 & 0.86 & 0.15 & 5.19 & 0.85 \\
6 & 0.73 & 0.78 & 0.15 & 5.52 & 0.80 \\
8 & 0.73 & 0.75 & 0.15 & 5.46 & 0.75 \\
\bottomrule
\end{tabular}
\end{table}

\begin{table}[t]
\caption{False-rejection analysis on randomized nominal trajectories without injected mission faults.}
\label{tab:false_rejection_analysis}
\centering
\begin{tabular}{ccccc}
\toprule
$H$ & FRR & Reject rate & Fallbacks & Mission succ. \\
\midrule
1 & 0.01 & 0.01 & 0.17 & 0.83 \\
2 & 0.02 & 0.05 & 0.67 & 0.67 \\
4 & 0.08 & 0.09 & 1.50 & 0.83 \\
6 & 0.11 & 0.11 & 1.50 & 0.67 \\
8 & 0.02 & 0.02 & 0.33 & 1.00 \\
\bottomrule
\end{tabular}
\end{table}

\begin{figure}[t]
\centering
\includegraphics[width=\columnwidth]{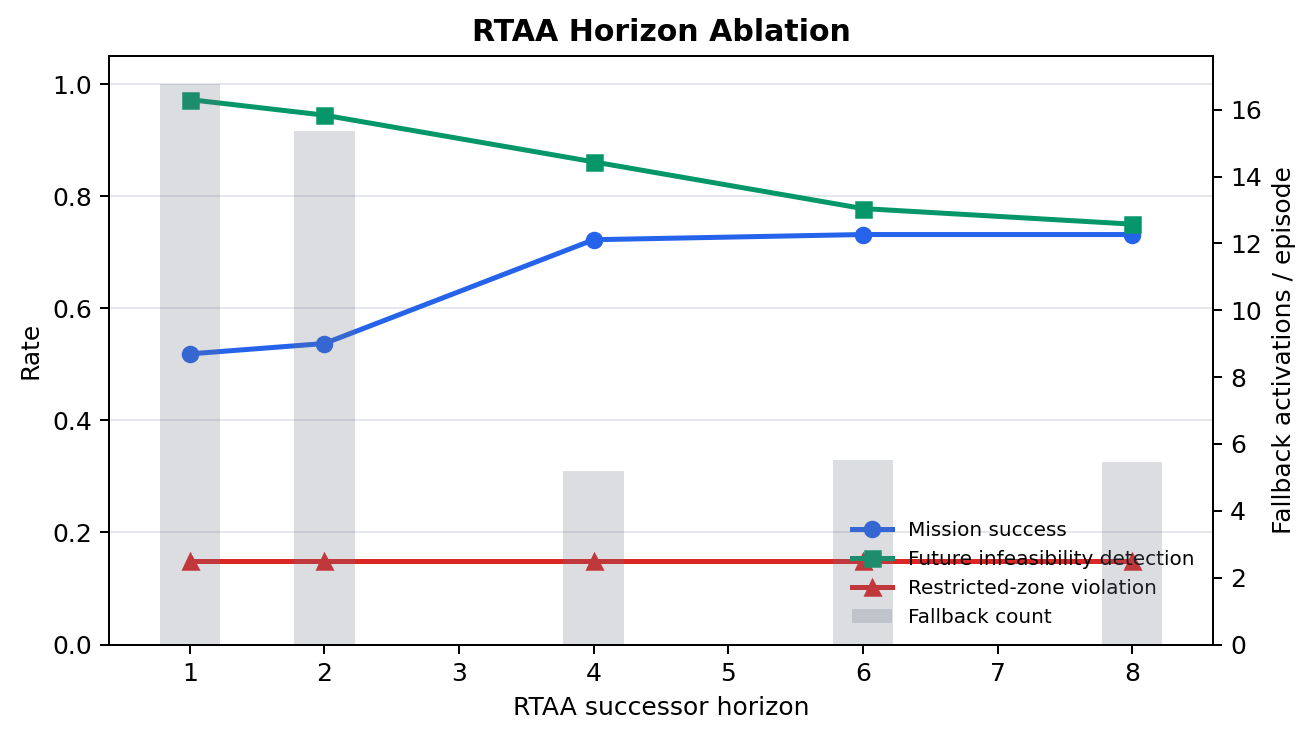}
\caption{Mission success and fallback activation under varying successor horizons.}
\label{fig:rtaa_horizon_ablation}
\end{figure}
\begin{figure*}[t]
\centering
\includegraphics[width=0.92\textwidth]{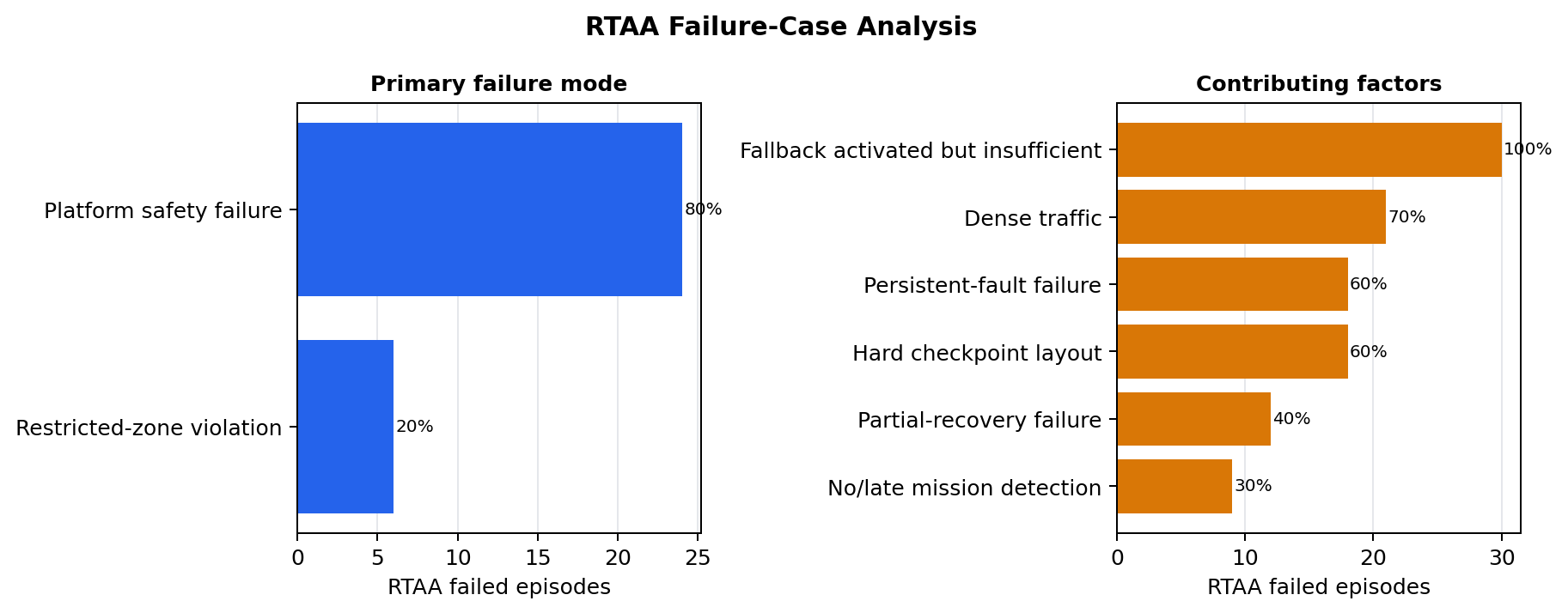}
\caption{Failure analysis under randomized mission-fault scenarios.}
\label{fig:rtaa_failure_analysis}
\end{figure*}
Table~\ref{tab:rtaa_horizon_ablation} and Fig.~\ref{fig:rtaa_horizon_ablation} show that short horizons react too late because they underestimate future mission infeasibility.
Horizons 1 and 2 therefore produce frequent fallback activation while still failing to preserve long-term mission feasibility.

As the successor horizon increases, the runtime monitor gains the ability to reject infeasible future evolution earlier before the system leaves the practical recoverable region.
However, longer horizons may also introduce more conservative intervention behavior.

In the current benchmark, $H=4$ provides the best practical tradeoff between mission success, fallback frequency, false rejection, and runtime overhead.

To further evaluate unnecessary intervention behavior, false rejection rate (FRR) is measured on nominal trajectories without injected faults:
\begin{equation}
\mathrm{FRR}
=
\frac{\text{rejected feasible commands}}
{\text{total feasible commands}}.
\end{equation}

Table~\ref{tab:false_rejection_analysis} shows that the proposed monitor maintains relatively low false rejection while still improving mission-fault detection.
The nonzero FRR indicates an expected tradeoff between early mission-infeasibility rejection and conservative runtime intervention.

\subsection{Failure Analysis}

Finally, we analyze the remaining RTAA failure cases under randomized mission-fault conditions.
The objective of this experiment is to determine whether the remaining failures originate from insufficient mission-infeasibility detection or insufficient fallback recovery capability.

\begin{table}[t]
\caption{Failure-case analysis for RTAA unsuccessful mission-fault episodes. Categories are not mutually exclusive for contributing factors.}
\label{tab:rtaa_failure_analysis}
\centering
\begin{tabular}{lcc}
\toprule
Failure category & Count & Share \\
\midrule
Platform safety failure & 24 & 0.80 \\
Restricted-zone violation & 6 & 0.20 \\
\midrule
Fallback activated but insufficient & 30 & 1.00 \\
Dense traffic & 21 & 0.70 \\
Persistent-fault failure & 18 & 0.60 \\
Hard checkpoint layout & 18 & 0.60 \\
Partial-recovery failure & 12 & 0.40 \\
No/late mission detection & 9 & 0.30 \\
\bottomrule
\end{tabular}
\end{table}

Table~\ref{tab:rtaa_failure_analysis} and Fig.~\ref{fig:rtaa_failure_analysis} show that most remaining failures occur after fallback activation has already been triggered.
This indicates that the proposed runtime monitor is generally capable of detecting mission-infeasible future evolution, but the current fallback controller is not always sufficiently capable of restoring mission feasibility under dense traffic or constrained checkpoint layouts.

The analysis further shows that dense traffic and difficult checkpoint configurations contribute to most unsuccessful RTAA episodes.
These scenarios reduce the practical recoverability region and limit the ability of the fallback controller to recover mission feasibility once inadmissible evolution has progressed too far.

The remaining failures therefore primarily reflect limitations in recovery authority rather than failures of mission-level fault detection itself.
These observations suggest that future work should combine mission-level runtime assurance with stronger recovery control and mission replanning mechanisms.

\section{Conclusion}

This paper presented a hierarchical mission-level runtime-assurance framework for autonomous driving under faulty high-level autonomy commands.
The central observation is that platform-safe behavior does not necessarily imply mission-feasible behavior.
The proposed successor-safe RTAA monitor validates candidate autonomy commands using future successor prediction before execution and rejects commands predicted to violate platform safety or mission feasibility.

Experiments in highway-env with mission-command fault injection show that conventional platform-level runtime-safety methods do not detect mission-level planning faults, while the proposed RTAA framework improves mission success under randomized fault conditions. The experimental results further demonstrate that runtime mission admissibility depends on both early future-infeasibility detection and effective fallback recovery. The proposed framework can interface with rule-based or learning-enabled autonomy modules through runtime command validation.
Future work will extend the framework to richer planning environments and integrate stronger recovery and mission-replanning strategies.

\bibliographystyle{IEEEtran}
\bibliography{references}

\end{document}